\documentclass[11pt]{article}
\usepackage{coling2020}
\usepackage{times}
\usepackage{url}
\usepackage{latexsym}
\usepackage[table,xcdraw]{xcolor}
\usepackage{graphicx}

\title{XED: \\A Multilingual Dataset for Sentiment Analysis and Emotion Detection}
  
\author{Emily Öhman \and Marc Pàmies \and Kaisla Kajava \and Jörg Tiedemann\\
University of Helsinki\\
{\tt firstname.lastname@helsinki.fi}}

\date{}

\begin{document}
\maketitle
\begin{abstract}
We introduce XED, a multilingual fine-grained emotion dataset. The dataset consists of human-annotated Finnish (25k) and English sentences (30k), as well as projected annotations for 30 additional languages, providing new resources for many low-resource languages. We use Plutchik's core emotions to annotate the dataset with the addition of neutral to create a multilabel multiclass dataset. The dataset is carefully evaluated using language-specific BERT models and SVMs to show that XED performs on par with other similar datasets and is therefore a useful tool for sentiment analysis and emotion detection. 
\end{abstract}

\section{Introduction}
There is an ever increasing need for labeled datasets for machine learning. This is true for English as well as other, often under-resourced, languages. We provide a cross-lingual fine-grained sentence-level emotion and sentiment dataset. The dataset consists of parallel manually annotated data for English and Finnish, with additional parallel datasets of varying sizes for a total of 32 languages created by annotation projection. We use Plutchik's Wheel of Emotions (\textit{anger}, \textit{anticipation}, \textit{disgust}, \textit{fear}, \textit{joy}, \textit{sadness}, \textit{surprise}, \textit{trust}) \cite{plutchik1980general} as our annotation scheme with the addition of \textit{neutral} on movie subtitle data from OPUS \cite{opus16}. 

We perform evaluations with fine-tuned cased multilingual and language specific BERT (Bidirectional Encoder Representations from Transformers) models \cite{Devlin_2019}, as well as Suport Vector Machines (SVMs). Our evaluations show that the human-annotated datasets behave on par with comparable state-of-the-art datasets such as the GoEmotions dataset \cite{demszky2020goemotions}. 
Furthermore, the projected datasets have accuracies that closely resemble human-annotated data with macro f1 scores of 0.51 for the human annotated Finnish data and 0.45 for the projected Finnish data when evaluating with FinBERT \cite{finbert}.

The XED dataset can be used in emotion classification tasks and other applications that can benefit from sentiment analysis and emotion detection such as offensive language identification. The data is open source\footnote{\url{https://github.com/Helsinki-NLP/XED}} licensed under a Creative Commons Attribution 4.0 International License (CC-BY).

In the following sections we discuss related work and describe our datasets. The datasets are then evaluated and the results discussed in the discussion section.

\blfootnote{
    %
    %
    \hspace{-0.65cm}  

    %
    This work is licensed under a Creative Commons 
    Attribution 4.0 International License.\\
    License details:
    \url{http://creativecommons.org/licenses/by/4.0/}.
}

\section{Background \& Previous Work}
Datasets created for sentiment analysis have been available for researchers since at least the early 2000s \cite{MANTYLA201816}. Such datasets generally use a binary or ternary annotation scheme (positive, negative + neutral) (e.g. \newcite{blitzer2007biographies})
and have traditionally been based on review data such as, e.g. Amazon product reviews, or movie reviews \cite{blitzer2007biographies,maas-EtAl:2011:ACL-HLT2011,Turney02}. Many, if not most, emotion datasets on the other hand use Twitter as a source and individual tweets as level of granularity \cite{schuff2017annotation,abdul2017emonet,mohammad-etal-2018-semeval}. In the case of emotion datasets, the emotion taxonomies used are often based on \newcite{ekman1971universals} and \newcite{plutchik1980general} (which is partially based on Ekman). 

\subsection{Existing Emotion Datasets}

\newcite{bostan2018analysis} analyze 14 existing emotion datasets of which only two are multilabel. These are AffectiveText \cite{strapparava2007semeval} and SSEC \cite{schuff2017annotation}. Nearly all of these datasets use an annotation scheme based on Ekman \cite{ekman1971universals,ekman1992argument} with many adding a few labels often following Plutchik's theory of emotions \cite{plutchik1980general}. A typical emotion dataset consists of 6-8 categories. The exception \newcite{bostan2018analysis} mention is CrowdFlower\footnote{CrowdFlower was created in 2016 but has since been acquired by different companies at least twice and is now hard to find. It is currently owned by Appen.} with 14 categories, and those not mentioned in Bostan et al. are e.g. the SemEval 2018 task 1 subtask c dataset \cite{mohammad-etal-2018-semeval} with 11 categories, EmoNet with 24 \cite{abdul2017emonet}, and the GoEmotions dataset \cite{demszky2020goemotions} with 27 categories. 

A majority of recent papers on multilabel emotion classification focus on the SemEval 2018 dataset which is based on tweets. Similarly, many of the non-multilabel classification papers use Twitter data. Twitter is a good base for emotion classification as tweets are limited in length and generally stand-alone, i.e. the reader or annotator does not need to guess the context in the majority of cases. Furthermore, hashtags and emojis are common, which further makes the emotion recognition easier for both human annotators and emotion detection and sentiment analysis models. Reddit data, as used by \newcite{demszky2020goemotions}, and movie subtitles used by this paper, are slightly more problematic as they are not "self-contained". Reddit comments are typically longer than one line and therefore provide some context for annotators to go by, but often lacks the hashtags and emojis of twitter and can be quite context-dependent as Reddit comments are by definition reactions to a post or another comment. Movie subtitles annotated out of sequence have virtually no context to aid the annotator and are supposed to be accompanied by visual cues as well. However, annotating with context can reduce the accuracy of one's model by doubly weighting surrounding units of granularity (roughly 'sentences' in our case) \cite{boland2013creating}. On the other hand, contextual annotations are less frustrating for the annotator and therefore, would likely provide more annotations in the same amount of time \cite{ohman2020challenges}.

In table \ref{datasetoverview} we have gathered some of the most significant emotion datasets in relation to this study. The table lists the paper in which the dataset was released (study), what the source data that was used was (source), what model was used to obtain the best evaluation scores (model), the number of categories used for annotation (cat), whether the system was multilabel or not (multi), and the macro f1 scores and accuracy score as reported by the paper (macro f1 and accuracy respectively). Some papers only reported a micro f1 and no macro f1 score. These scores have been marked with a $\mu$.

\begin{table}[htbp!]
\resizebox{\textwidth}{!}{
\begin{tabular}{lllllll}
\rowcolor[HTML]{EFEFEF} 
study                            & source              & model              & cat        & multi      & macro f1         & accuracy  \\
\newcite{strapparava2007semeval} & news headlines      & "CLaC system"      & 6+val      & yes        & N/A                & 55.1\%    \\
\newcite{tokuhisa2008emotion}    & JP web corpus       & k-NN               & 10         & no         & N/A              & N/A       \\
\newcite{schuff2017annotation}   & Twitter             & BiLSTM             & 8          & yes        & N/A              & N/A       \\
\newcite{abdul2017emonet}        & Twitter             & GRNNs              & 6          & no         & N/A              & 95.68\%   \\
\newcite{abdul2017emonet}        & Twitter             & GRNNs              & 24         & no         & 87.47\%          & N/A       \\
\newcite{SAMY201861}             & Twitter             & C-GRU              & 11+neu     & yes        & 64.8\%           & 53.2\%    \\
\newcite{yu-etal-2018-improving} & Twitter             & DATN               & 11         & yes        & 44.4\%           & 45.7\%    \\
\newcite{jabreel}                & Twitter             & BiRNN/GRU          & 11         & yes        & 56.4\%           & 59\%      \\
\newcite{huang2019seq2emo}       & Twitter             & BiLSTM/ELMO        & 11         & yes        & 70.9\%$\mu$      & 59.2\%    \\
\newcite{Liu2019DENSAD}          & books etc           & BERT               & 8+neu      & no         & 60.4\%$\mu$      & N/A         \\
\newcite{demszky2020goemotions}  & Reddit              & BERT               & 27         & yes        & 46\%             & N/A       \\
\newcite{demszky2020goemotions}  & Reddit              & BERT               & 6          & no        & 64\%             & N/A       \\
XED (English)                    & movie subtitles     & BERT               & 8+neu      & yes        & 53.6\%           & 54.4\%     
\end{tabular}}
\caption{Overview of emotion datasets.}
\label{datasetoverview}
\end{table}


The datasets in table \ref{datasetoverview} differ from each so much in content, structure, and manner of annotation that direct comparisons are hard to make. Typically, the fewer the number of categories, the easier the classification task and the higher the evaluation scores. It stands to reason that the easier it is to detect emotions in the source data, the easier it is for annotators to identify and agree upon annotation labels and therefore it becomes easier for the system or model to correctly classify the test data as well. The outlier in these datasets is EmoNet \cite{abdul2017emonet} which achieved astonishing accuracies by using 665 different hashtags to automatically categorize 1.6 million tweets into 24 categories (Plutchik's 8 at 3 different intensities), unfortunately neither the dataset or their model has been made available for closer inspection. 

The downside of datasets trained on Twitter is that they are likely not that good at classifying anything other than tweets. It is plausible that datasets trained on less specific data such as XED and those created by \newcite{tokuhisa2008emotion} and \newcite{demszky2020goemotions} are better at crossing domains at the cost of evaluation metrics.
\subsection{Annotation Projection}
Research shows that affect categories are quite universal \cite{Cowen2019ThePO,Scherer1994EvidenceFU}. Therefore, theoretically they should also to a large degree retain emotion categories when translated.
Annotation projection has been shown to offer reliable results in different NLP and NLU tasks \cite{kajava2020emotion,yarowsky2001inducing,agic2016multilingual,rasooli-tetrault-2015}. Projection is sometimes the only feasible way to produce resources for under-resourced languages. By taking datasets created for high-resource languages and projecting these results on the corresponding items in the under-resourced language using parallel corpora, we can create datasets in as many languages as exist in the parallel corpus. A parallel corpus for multiple languages enables the simultaneous creation of resources for multiple languages at a low cost.


Previous annotation tasks have shown that even with binary or ternary classification schemes, human annotators agree only about 70-80\% of the time and the more categories there are, the harder it becomes for annotators to agree \cite{boland2013creating,mozetivc2016multilingual}. For example, when creating the DENS dataset \cite{Liu2019DENSAD}, only 21\% of their annotations had consensus between all annotators with 73.5\% having to resort to majority agreement, and a further 5.5\% could not be agreed upon and were left to expert annotators to be resolved. 

Some emotions are also harder to detect, even for humans. \newcite{demszky2020goemotions} show that the emotions of \textit{admiration, approval, annoyance, gratitude} had the highest interrater correlations at around 0.6, and \textit{grief, relief, pride, nervousness, embarrassment} had the lowest interrater correlations between 0-0.2, with a vast majority of emotions falling in the range of 0.3-0.5 for interrater correlation. 
Emotions are also expressed differently in text with \textit{anger} and \textit{disgust} expressed explicitly, and \textit{surprise} in context \cite{Alm2005EmotionsFT}.

Some emotions are also more closely correlated. In Plutchik's wheel \cite{plutchik1980general} related emotions are placed on the same dyad so that for example for \textit{anger} as a core emotion, there is also \textit{rage} that is more intense, but highly correlated with anger, and \textit{annoyance} which is less intense, but equally correlated. In this way it is also possible to map more distinct categories of emotions onto larger wholes; in this case \textit{rage} and \textit{annoyance} could be mapped to \textit{anger}, or even more coarsely to \textit{negative}. 
This approach has been employed by for example \newcite{abdul2017emonet}.
\section{The Data}
In table \ref{tab:quicklook} we present an overview of the English part of the XED dataset. With 24,164 emotion annotations\footnote{Note that the total number of annotations excluding \textit{neutral} (24,164) and the combined number of annotations (22,424) differ because once the dataset was saved as a Python dictionary, identical lines were merged as one (i.e. some common movie lines like "All right then!" and "I love you" appeared multiple times from different sources).} excluding \textit{neutral} on 17,520 unique sentences\footnote{A sentence could have been annotated as containing 3 different emotions by one or more annotators. This would count as 3 annotations on one unique data point.} the XED dataset is one of the largest emotion datasets we are aware of.


We used Plutchik's core emotions as our annotation scheme resulting in 8 distinct emotion categories plus neutral. The \textit{Sentimentator} platform \cite{ohman-senti2018,ohman2018creating} allows for the annotation of intensities resulting in what is essentially 30 emotions and sentiments, however, as the intensity score is not available for all annotations, the intensity scores were discarded. The granularity of our annotations roughly correspond to sentence-level annotations, although as our source data is movie subtitles, our shortest subtitle is \textit{!} and the longest subtitle consists of three separate sentences.

\begin{table}[htbp!]
\centering
\begin{tabular}{|
>{\columncolor[HTML]{EFEFEF}}l r|}
\hline
Number of annotations:           & 24,164 + 9,384 neutral                                                                         \\ \hline
Number of unique data points:    & 17,520 + 6,420 neutral                                                                           \\ \hline
Number of emotions:              & 8 (+pos, neg, neu)                                                                  \\ \hline
Number of annotators:            & 108 (63 active)                                                                \\ \hline
Number of labels per data point: & \begin{tabular}[c]{@{}r@{}}1:\ \ \ 78\%\\ 2:\ \ \ 17\%\\ 3:\ \ \ \ \ 4\%\\ 4+: 0.8\%\end{tabular} \\ \hline
\end{tabular}
\caption{Overview of the XED English dataset.}
\label{tab:quicklook}
\end{table}

A majority of the subtitles for English were assigned one emotion label (78\%), 17\% were assigned two, and roughly 5\% had three or more categories (see also Table 3). 


\subsection{Movie Subtitles as Multilingual Multi-Domain Proxy}
We use the OPUS \cite{opus16} parallel movie subtitle corpus of subtitles collected from opensubtitles.org as a multi-domain proxy. As the movies we use for source data cover several different genres and, although scripted, represents real human language used in a multitude of situations similar to many social media platforms.

Because OPUS open subtitles is a parallel corpus we are able to evaluate our annotated datasets across languages and at identical levels of granularity. Although the subtitles might be translated using different translation philosophies (favoring e.g. meaning, mood, or idiomatic language as the prime objective) \cite{carl2011taxonomy}, we expect the translations to have aimed at capturing the sentiments and emotions originally expressed in the film based on previous studies (e.g. \newcite{Cowen2019ThePO}, \newcite{Scherer1994EvidenceFU}, \newcite{creutz}, \newcite{scherrer2020tapaco} and \newcite{kajava2020emotion}). 

\subsection{Data Annotation}
The vast majority of the dataset was annotated by university students learning about sentiment analysis with some annotations provided by expert annotators for reliability measurements \cite{ohman2018creating}. 
The students' annotation process was monitored and evaluated. They received only minimal instructions. These instructions included that they were to focus on the quality of annotations rather than quantity, and to annotate from the point of view of the speaker. We also asked for feedback on the annotation process to improve the user-friendliness of the platform for future use. In tables \ref{tab:quicklook} and \ref{tab:FIquicklook} the number of active annotators have been included. All in all over 100 students annotated at least some sentences with around 60 active annotators, meaning students who annotated more than 300 sentences \cite{ohman2020challenges}.

It should be noted that the annotators were instructed to annotate the subtitles without context, a task made harder by the fact that we chose subtitles that were available for all languages, which likely meant that some of the most famous movies were included thus creating recognizable context for the annotators.

The data for annotation was chosen randomly from the OPUS subtitle corpus \cite{opus16} from subtitles that were available for the maximum number of languages. We chose 30,000 individual lines to be annotated by 3 annotators. For the final dataset, some of these annotations were not annotated by all 3 annotators, as it was possible to skip difficult-to-annotate instances, but the subtitle was included if at least 2 annotators agreed on the emotion score. In some cases if the expert annotators agreed that the annotation was feasible during the pre-processing phase, subtitles annotated by a single annotator and checked by expert annotators, were also included.

\subsection{Pre-processing}
After the annotations were extracted from the database, the data needed to be cleaned up. The different evaluations required different pre-processing steps. Most commonly, this included the removal of superfluous characters containing no information. We tried to keep as much of the original information as possible, including keeping offensive, racist, and sexist language as is. If such information is removed, the usefulness of the data is at risk of being reduced, particularly when used for e.g. offensive language detection \cite{LT@offenseval}.

For the English data we used Stanford NER (named entity recognition) \cite{stanfordNER} to replace names and locations with the tags: [PERSON] and [LOCATION] respectively. We kept organization names as is because we felt that the emotions and sentiments towards some large well-known organizations differ too much (cf. IRS, FBI, WHO, EU, and MIT). For the Finnish data, we replaced names and locations using the Turku NER corpus \cite{luoma2020broad}.


Some minor text cleanup was also conducted, removing hyphens and quotations marks, and correcting erroneous renderings of characters (usually encoding issues) where possible.

\subsection{English Dataset Description}
The final dataset contained 17,520 unique emotion-annotated subtitles as shown in table \ref{xeddescr}. In addition there are some 6.5k subtitles annotated as \textit{neutral}. The label distribution can be seen in table \ref{xeddescr}.

\begin{table}[htbp!]
\resizebox{\textwidth}{!}{
\begin{tabular}{lllllllll}
\rowcolor[HTML]{EFEFEF} 
anger                          & anticipation                   & disgust                        & fear                           & joy                           & sadness                        & surprise                      & trust                          & Total annotations                                                \\
4,182                           & 3,660                           & 2,442                           & 2,585                           & 3,139                          & 2,635                           & 2,635                          & 2,886                           & \cellcolor[HTML]{EFEFEF}24,164                                    \\
17.31\%                        & 15.15\%                        & 10.11\%                        & 10.70\%                        & 12.99\%                       & 10.90\%                        & 10.90\%                       & 11.94\%                        & \cellcolor[HTML]{EFEFEF}XED percentage                           \\
{\color[HTML]{9B9B9B} 15.09\%} & {\color[HTML]{9B9B9B} 10.15\%} & {\color[HTML]{9B9B9B} 12.80\%} & {\color[HTML]{9B9B9B} 17.86\%} & {\color[HTML]{9B9B9B} 8.34\%} & {\color[HTML]{9B9B9B} 14.41\%} & {\color[HTML]{9B9B9B} 6.46\%} & {\color[HTML]{9B9B9B} 14.89\%} & \cellcolor[HTML]{EFEFEF}{\color[HTML]{9B9B9B} EmoLex percentage}
\end{tabular}}
\caption{Emotion label distribution in the XED English dataset.}
\label{xeddescr}
\end{table}
The emotion labels are surprisingly balanced with the exception of \textit{anger} and \textit{anticipation}, which are more common than the other labels. In comparison with one of the most well-known emotion datasets using the same annotation scheme, the NRC emotion lexicon (EmoLex) \cite{Mohammad13}, the distribution differs somewhat. Although \textit{anger} is a large category in both datasets, \textit{fear} is average in our dataset, but the largest category in EmoLex. It is hard to speculate why this is, but one possible reason is the different source data. 


The number of unique label combinations is 147, including single-label. The most common label combinations beyond single-label are \textit{anger} with \textit{disgust} (2.4\%) and \textit{joy} with \textit{trust} (2.1\%) followed by different combinations of the positive emotions of \textit{anticipation}, \textit{joy}, and \textit{trust}. These findings are in line with previous findings discussing overlapping categories \cite{Banea11multilingualsentiment,demszky2020goemotions}. However, these are followed by \textit{anger} combined with \textit{anticipation} and \textit{sadness} with \textit{surprise}. The first combination is possibly a reflection of the genre, as a common theme for \textit{anger} with \textit{anticipation} is threats. The combination of \textit{surprise} with negative emotions (\textit{anger, disgust, fear, sadness}) is much more common than a combination with positive emotions.

Note that the difference between total annotations excluding \textit{neutral} (24,164) and the combined number of annotations (22,424) differ because once the dataset was saved as a Python dictionary, identical lines were merged as one (i.e. some common movie lines like "All right then!" and "I love you" appeared multiple times from different sources). Additionally, lines annotated as both \textit{neutral} and an emotion were removed from the \textit{neutral} set.

\subsubsection{Crosslingual Data \& Annotation projection}

From our source data we can extract parallel sentences for 43 languages. For 12 of these languages we have over 10,000 sentences available for projection as per table \ref{tab:pairs}. We removed some of these languages for having fewer than 950 lines, resulting in a total of 32 languages\footnote{Arabic (AR), Bosnian (BS), Brazilian Portuguese (PT\_BR), Bulgarian (BG), Croatian (HR), Czech (CS), Danish (DA), Dutch (NL), English (EN), Estonian (ET), Finnish (FI), French (FR), German (DE), Greek (EL), Hebrew (HE), Hungarian (HU), Icelandic (IS), Italian (IT), Macedonian (MK), Norwegian (NO), Polish (PL), Portuguese (PT), Romanian (RO), Russian (RU), Serbian (SR), Slovak (SK), Slovenian (SL), Spanish (ES), Swedish (SV), Turkish (TR) and Vietnamese (VI)} including the annotated English and Finnish data. We have made all 32 datasets available on GitHub plus the raw data for all 43 languages including the 11 datasets that had fewer than 950 lines.

\begin{table}[htbp!]
\centering
\resizebox{\textwidth}{!}{
\begin{tabular}{ccccccccccccc}
\rowcolor[HTML]{EFEFEF} 
IT    & FI    & FR    & CS    & PT    & PL    & SR    & TR    & EL    & RO    & ES       & PT\_BR \\
10,582 & 11,128 & 11,503 & 11,885 & 12,559 & 12,836 & 14,831 & 15,712 & 15,713 & 16,217 & 16,608 & 22,194
\end{tabular}}
\caption{Languages (ISO code) with over 10k parallel sentences with our annotated  English data.}
\label{tab:pairs}
\end{table}

To test how well our data is suited for emotion projection, we projected the English annotations onto our Finnish unannotated data using OPUS tools \cite{aulamo2020opustools}. We chose Finnish as our main test language as we also have some annotated data for it to use as a test set. The manually annotated Finnish data consists of nearly 20k individual annotations and almost 15k unique annotated sentences plus an additional 7,536 sentences annotated as \textit{neutral} \footnote{The same calculations apply here as for English. Annotations are calculated as labels which can be more than one for each line, and unique data points refer to the number of lines that had 1 or more annotations.}. The criteria for the inclusion of an annotation was the same as for English. The distribution of the number of labels and the labels themselves are quite similar to that of the English data. Relatively speaking there is a little less \textit{anticipation} in the Finnish data, but \textit{anger} is the biggest category in both languages. 

\begin{table}[htbp!]
\centering
\begin{tabular}{|
>{\columncolor[HTML]{EFEFEF}}l r|}
\hline
Number of annotations:           & 21,984 (/w neutral)                                                                          \\ \hline
Number of unique data points:    & 14,449 + 7,536 neutral                                                                          \\ \hline
Number of emotions:              & 8 (+neu)                                                                  \\ \hline
Number of annotators:            & ~40 active                                                               \\ \hline
Number of labels per data point: & \begin{tabular}[c]{@{}r@{}}1:\ \ \ 77\%\\ 2:\ \ \ 18\%\\ 3: \ \ \ 5\%\\ 4+: 1.2\%\end{tabular} \\ \hline
\end{tabular}
\caption{Overview of the XED Finnish dataset.}
\label{tab:FIquicklook}
\end{table}

The distribution of the number of emotions (table \ref{tab:FIquicklook}) and the distribution of emotions (table \ref{FIxed}) are similar to their corresponding distributions in the English dataset.

\begin{table}[htbp!]
\centering
\begin{tabular}{lllllllll}
\rowcolor[HTML]{EFEFEF} 
anger  & anticipation & disgust & fear   & joy    & sadness & surprise & trust  & Total annotations      \\
3,345   & 2,496         & 2,373    & 2,186   & 2,559   & 2,184    & 1,982     & 2,404   & 19,529 \\
17.13\% & 12.78\%       & 12.15\%  & 11.19\% & 13.10\% & 11.18\%  & 10.15\%   & 12.31\% & 100\%     
\end{tabular}
\caption{Emotion label distribution in the XED Finnish dataset.}
\label{FIxed}
\end{table}

We used the 11,128 Finnish sentences for which directly parallel sentences existed and projected the English annotations on them using the unique alignment IDs for both languages as guide. Some of those parallel sentences were part of our already annotated data and were discarded as training data. This served as a useful point of comparison. The average annotation correlation using Cohen's kappa is 0.44 (although accuracy by percentage is over 90\%), and highest for \textit{joy} at 0.65, showing that annotation projection differs from human annotation to a similar degree as human annotations differ from each other.




\section{Evaluation}
A dataset for classification tasks is useful only if the accuracy of its annotations can be confirmed. To this end we use BERT to evaluate our annotations as it has consistently outperformed other models in recent classification tasks (see e.g \newcite{zampieri2020semeval}), and Support Vector Machines for its simplicity and effectiveness. We use a stratified split of 70:20:10 for training, dev, and test data.

We use a fine-tuned English uncased BERT, with a batch size of 96. The learning rate of Adam optimizer was set to 2e-5 and the model was trained for 3 epochs. The sequence length was set to 48. We perform a 5-fold cross validation.

We also use an SVM classifier with linear kernel and regularization parameter of 1. Word unigrams, bigrams and trigrams were used as features in this case. Implementation was done using the LinearSVC class from the scikit-learn library \cite{scikit}.

\textit{Binary} refers to \textit{positive} and \textit{negative}, and \textit{ternary} refers to \textit{positive, negative}and \textit{neutral}. For binary evaluations we categorized \textit{anger}, \textit{disgust}, \textit{fear}, and \textit{sadness} as \textit{negative}, and \textit{anticipation}, \textit{joy}, and \textit{trust} as \textit{positive}. \textit{Surprise} was either discarded or included as a separate category (see table \ref{tab:Eneval}). For this classification task BERT achieved macro f1 scores of 0.536 and accuracies of 0.544. This is comparable to other similar datasets when classes are merged (e.g. \newcite{demszky2020goemotions}).

\subsection{Evaluation Metrics}
We achieve macro f1 scores of 0.54 for our multilabel classification with a fine-tuned BERT model. Using named-entity recognition increases the accuracy slightly. For binary data mapped from the emotion classifications onto positive and negative (non-multilabel classification) our model achieves a macro f1 score of 0.838 and accuracy of 0.840. Our linear SVM classifier using one-vs-rest achieves an f1 score of 0.502 with per class f1 scores between 0.8073 (anger) and 0.8832 (fear \& trust) (see tables \ref{tab:Eneval} and \ref{tab:SVCpercalss}).

\begin{table}[htbp!]
\parbox{.5\linewidth}{
\centering
\small
\begin{tabular}{@{}lll}
\rowcolor[HTML]{EFEFEF} 
data                                   & f1    & accuracy \\
English without NER, BERT              & 0.530 & 0.538    \\
English with NER, BERT                 & 0.536 & 0.544    \\
English NER with neutral, BERT         & 0.467 & 0.529    \\
English NER binary with surprise, BERT & 0.679 & 0.765    \\
English NER true binary, BERT          & 0.838 & 0.840    \\
Finnish anno., FinBERT & 0.507 & 0.513 \\
English NER, one-vs-rest SVM (LinearSVC)\footnotemark    & 0.746 &

\end{tabular}
\caption{Evaluation results of the XED English dataset.}
\label{tab:Eneval}
}
\parbox{.6\linewidth}{
\centering
\small
\begin{tabular}{@{}ll}
\rowcolor[HTML]{EFEFEF} 
SVM per class f1 & emotion      \\
0.8073           & anger        \\
0.8296           & anticipation \\
0.8832           & disgust      \\
0.8763           & fear         \\
0.8819           & joy          \\
0.8762           & sadness      \\
0.8430           & surprise     \\
0.8832           & trust        

\end{tabular}
\caption{SVM per class f1 scores.}
\label{tab:SVCpercalss}}
\end{table}
\footnotetext{The SVM evaluation was performed on the student annotations only in order to be fully comparable to the projections. The BERT evaluations also contain additional data from the Sentimentator and cynarr GitHub repos. These are linked to from the main XED repo.}

\begin{figure}[htbp!]
\centerline{\includegraphics[scale=.65]{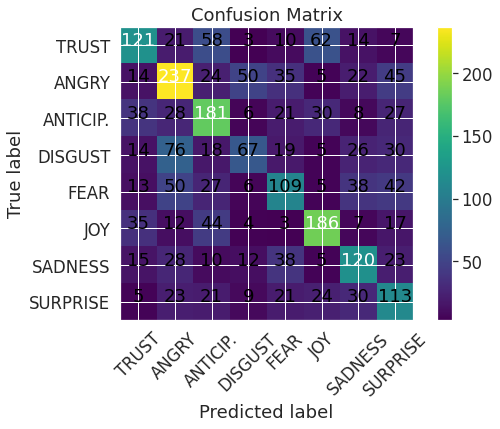}}
\caption{Confusion matrix for the XED English dataset.}
\label{fig:cm}
\end{figure}
\begin{figure}[htbp!]
\centerline{\includegraphics[scale=1]{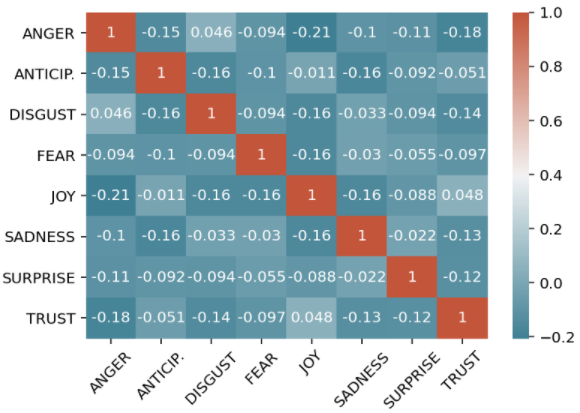}}
\caption{Correlation matrix for the XED English dataset.}
\label{fig:corrm}
\end{figure}

The confusion matrix (see Figure \ref{fig:cm}) reveals that \textit{disgust} is often confused with \textit{anger}, and to some extent this is true in the other direction as well. This relation between labels can also be observed in the correlation matrix (see Figure \ref{fig:corrm}), where \textit{anger} and \textit{disgust} appear as one of the most highly correlated pair of categories, only behind \textit{joy} and \textit{trust}. On the other hand, the least correlated pair is \textit{joy} and \textit{anger}, closely followed by \textit{trust} and \textit{anger}. \textit{Disgust} is also the hardest emotion to categorize correctly. In fact, it is more often classified as \textit{anger} than \textit{disgust}. \textit{Joy, anger} and \textit{anticipation} are the categories that are categorized correctly the most.

The correlation matrix for the multilabel English evaluation shows (table \ref{fig:corrm}) how closely correlated the emotions of \textit{anger} and \textit{disgust}, and \textit{joy} and \textit{trust} in particular are.

\subsection{Evaluating Annotation Projection}

With the same parameters as for English, we used language-specific BERT models from Huggingface transformers \cite{huggingface} for the Arabic, Chinese, Dutch, Finnish, German and Turkish datasets with 5-fold cross-validation. The annotated Finnish dataset achieves an f1 score of 0.51. The projected annotations achieve slightly worse f1 scores than the annotated dataset at 0.45 for Finnish (see table \ref{tab:projectedeval}). The other datasets achieve similar f1 scores, with the Germanic languages of German and Dutch achieving almost as high scores as the original English dataset. This is likely a reflection of typological, cultural, and linguistic similarities between the languages making the translation to begin with more similar to the original and therefore minimizing information loss.
\begin{table}[htbp!]
\centering
\small
\begin{tabular}{lll}
\rowcolor[HTML]{EFEFEF} 
data                 & f1 & accuracy \\

Finnish projected & 0.4461 &0.4542\\
Turkish projected & 0.4685 &0.5257\\
Arabic projected & 0.4627 &0.5339\\
German projected &0.5084&0.5737\\
Dutch projected&0.5155 & 0.5822\\
Chinese projected&0.4729&0.5247
\end{tabular}
\caption{Annotation projection evaluation results using language-specific BERT models.}
\label{tab:projectedeval}
\end{table}

We also evaluated all the projected datasets using a linear SVC classifier. In most cases the linear SVC classifier performs better than language-specific BERT. We speculate this is related to the size of the datasets.
\begin{table}[htbp!]
\centering
\begin{tabular}{ccccccccccc}
\rowcolor[HTML]{EFEFEF} 
AR      & BG     & BS     & CN     & CS     & DA     & DE     & EL     & ES     & ET     & \\
0.5729  & 0.6069 & 0.5854 & 0.5004 & 0.6263 & 0.5989 & 0.6059 & 0.6192 & 0.6760 & 0.5449 & \\
\rowcolor[HTML]{EFEFEF} 
FI     & FR     & HE     & HR      & HU     & IS     & IT     & MK     & NL     & NO     & \\
0.5859 & 0.6257 & 0.5980 & 0.6503  & 0.5978 & 0.5416 & 0.6907 & 0.4961 & 0.6140 & 0.5771 & \\
\rowcolor[HTML]{EFEFEF} 
PL      & PT     & PT\_BR & RO     & RU     & SK     & SL     & SR     & SV     & TR      & VI \\
0.6233  & 0.6203 & 0.6726 & 0.6387 & 0.6976 & 0.5305 & 0.6015 & 0.6566 & 0.6218 & 0.6080  & 0.5594 \\
\end{tabular}
\caption{SVM's macro f1 scores for all projected languages.}
\label{tab:all-evals}
\end{table}

\section{Discussion}
The results from the dataset evaluations show that the XED is on par with other similar datasets, but they also stress that reliable emotion detection is still a very challenging task. It is not necessarily an issue with natural language processing and understanding as these types of tasks are challenging for human annotators alike. If human annotators cannot agree on labels, it is not reasonable to think computers can do any better regardless of annotation scheme or model used since these models are restricted by human performance. The best accuracies are those that are in line with annotator agreement. 

XED is a novel state-of-the-art dataset that provides a new challenge in fine-grained emotion detection with previously unavailable language coverage. What makes the XED dataset particularly valuable is the large number of annotations at high granularity, as most other similar datasets are annotated at a much coarser granularity. The use of movie subtitles as source data means that it is possible to use the XED dataset across multiple domains (e.g. social media) as the source data is representative of other domains and not as restricted to the domain of the source data (movies) as many other datasets. Perhaps the greatest contribution of all is that, for the first time, many under-resourced languages have emotion datasets that can be used in other possible downstream applications as well.

\bibliographystyle{acl}
\bibliography{dataset}

\end{document}